\documentclass[letterpaper, 10pt, conference]{ieeeconf}      
\IEEEoverridecommandlockouts
\usepackage{graphicx}
\usepackage{subcaption}
\usepackage{amsmath,amssymb}
\usepackage{algorithmic}
\usepackage{algorithm}
\usepackage{xcolor}
\usepackage{tabularx}
\usepackage{booktabs}
\usepackage{multirow}
\usepackage{orcidlink}

\usepackage[nolist,nohyperlinks]{acronym}
\usepackage{hyperref}
\hypersetup{
    colorlinks=true,
    linkcolor=blue,
    citecolor=teal,
    filecolor=magenta,      
    urlcolor=cyan,
    pdftitle={main},
    pdfpagemode=FullScreen,
    }
\usepackage[backend=biber,style=ieee,sorting=none,abbreviate=true]{biblatex}

\begin{acronym}
    \acro{IHM}{In-Hand Manipulation}
    \acro{CL}{Continual Learning}
    \acro{RL}{Reinforcement Learning}
    \acro{CPD}{Continual Policy Distillation}
    \acro{PPO}{Proximal Policy Optimization}
    \acro{PD}{Policy Distillation}
    \acro{CPD}{Continual Policy Distillation}
    \acro{ID}{Identifier}
\end{acronym}

\title{\LARGE \bf Continual Policy Distillation of Reinforcement Learning-based\\Controllers for Soft Robotic In-Hand Manipulation}
\author{Lanpei Li$^{*,1}$ \orcidlink{0009-0005-4370-1020}, Enrico Donato$^{*,2}$ \orcidlink{0000-0002-8844-5279}, Vincenzo Lomonaco$^{\star,1}$ \orcidlink{0000-0001-8308-6599}, Egidio Falotico$^{\star,2}$ \orcidlink{0000-0001-8060-8080}
\thanks{$*$ These authors equally contributed to the work.}
\thanks{$\star$ These authors equally supervised the work.}
\thanks{This work received funding from the European Union’s Horizon 2020 research and innovation program under grant agreement No. 863212 (PROBOSCIS project) and from the EBRAINS (European Brain Research Infrastructures Italy) project under Grant B51E22000150006.}
\thanks{$^{1}$ L. Li and V. Lomonaco are with the Department of Computer Science, University of Pisa, Pisa, 56127, Italy. L. Li also holds an affiliation with the Institute of Information Science and Technologies (ISTI), National Research Council (CNR), Pisa, 56124, Italy. {\tt\small lanpei.li@phd.unipi.it, vincenzo.lomonaco@unipi.it}}%
\thanks{$^{2}$E. Donato and E. Falotico are with The BioRobotics Institute, Sant'Anna School of Advanced Studies, 56025 Pontedera (PI), Italy and with the Departement of Excellence in Robotics \& AI, Sant'Anna School of Advanced Studies, 56125 Pisa, Italy {\tt\small \{e.donato, e.falotico\}@santannapisa.it}}%
}

\begin{document}
    \maketitle   
    
    \begin{abstract}
        Dexterous manipulation, often facilitated by multi-fingered robotic hands, holds solid impact for real-world applications. Soft robotic hands, due to their compliant nature, offer flexibility and adaptability during object grasping and manipulation. Yet, benefits come with challenges, particularly in the control development for finger coordination. Reinforcement Learning (RL) can be employed to train object-specific in-hand manipulation policies, but limiting adaptability and generalizability. We introduce a Continual Policy Distillation (CPD) framework to acquire a versatile controller for in-hand manipulation, to rotate different objects in shape and size within a four-fingered soft gripper. The framework leverages Policy Distillation (PD) to transfer knowledge from expert policies to a continually evolving student policy network. Exemplar-based rehearsal methods are then integrated to mitigate catastrophic forgetting and enhance generalization. The performance of the CPD framework over various replay strategies demonstrates its effectiveness in consolidating knowledge from multiple experts and achieving versatile and adaptive behaviours for in-hand manipulation tasks. 
    \end{abstract}

    
    \section{Introduction}
        Dexterous manipulation is a crucial and essential skill for biological systems for everyday interaction tasks. In-hand manipulation encompasses grasping, rotating, and translating objects within the hand, closely emulating the dexterity exhibited by humans. It has garnered significant attention within the robotic research community \cite{billard2019manipulation}, driven by its far-reaching implications for real-world applications. Such interest has, in turn, led to deep scientific investigations.
        
        Successful manipulation hinges upon two fundamental prerequisites: interaction facilitation and precise movement coordination \cite{villani2012controlhands}. To this aim, a significant effort has been dedicated to emulating the capabilities of human hands, such as the integration of multi-fingered grippers \cite{piazza2019hands} to enable more dexterous interactions. Last achievements show how soft robotic hands \cite{shintake2018softgrippers} have garnered more attention due to their advantages in grasping and manipulating unfamiliar objects. The compliant nature of soft robot hands renders them more adaptable to complex and unpredictable environments, by mimicking the dexterity and flexibility of human hands.

        These endeavours have not come without their challenges. The management of these advanced artefacts has seen a notable rise in intricacy concerning precise position control. This surge can be traced back to the expansion in degrees of freedom, alongside the necessity for refined finger coordination \cite{villani2012controlhands}. Furthermore, the task becomes even more demanding due to the need to maintain contact while averting slippage \cite{stachowsky2016slippage}. Nevertheless, soft manipulation capitalizes on compliant behaviour to navigate such obstacles efficiently, leading to robust and adaptable manipulation \cite{puhlmann2022softhand}. 

        \begin{figure}[t]
            \includegraphics[width=\linewidth]{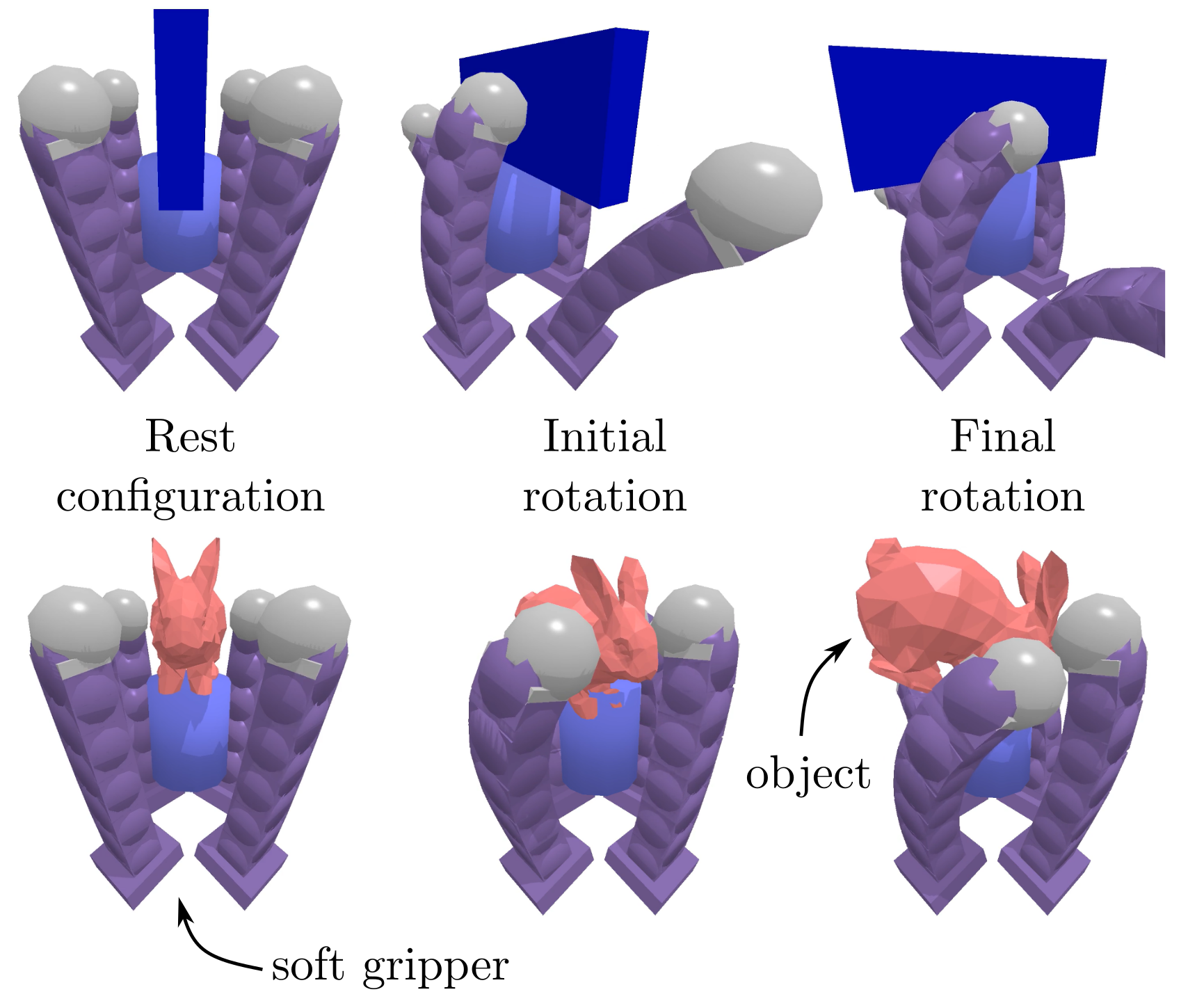}
            \captionsetup{justification=centering}
            \caption{Simulation of In-hand Manipulation}
            \label{fig:simulators}
        \end{figure} 

        AI-based algorithms have demonstrated their efficiency in learning control policies for robotic in-hand manipulation \cite{kroemer2021learningmanipulation}, typically resulting in specific object controllers without retaining the training environment or data. However, the challenge arises when dealing with multiple object-specific controllers, since the development of a single, versatile controller for general-purpose manipulation is still under debate. Real-world applications often face limitations in accessing specific control policies sequentially, constraining traditional batch learning paradigms \cite{cossu2021clrnn}. These challenges raise fundamental questions about how to effectively utilize previously acquired policies for subsequent training and implement continually learning agents to enhance their capabilities progressively \cite{lesort2020clrobotics}. In the context of soft robotics, where both the action space (actuation) and observation space (sensing) are continuous, learning manipulation tasks becomes even more complex, and training \ac{RL}-based controllers for multiple objects simultaneously is hindered by interactive learning, computational, and memory constraints. Additionally, traditional replay-based approaches may compromise data privacy by storing raw or processed data that could contain sensitive information.
        
        Our contribution aims at overcoming such limitations and proposes the implementation of the \ac{CPD} framework to generate experts' demonstrations for object-specific in-hand manipulation and their use to continually train a further expert sequentially and asynchronously. Such approach ensures that the previously acquired knowledge is retained and not lost throughout the process. The algorithm is implemented on a four-fingered soft gripper for in-hand rotation of objects with variable shapes, demanding for precise finger coordination as shown in Fig. \ref{fig:simulators}. We evaluate the algorithm's performance, with particular emphasis on offline demonstration sampling, the buffer memory size for the retention of previously acquired knowledge, and the criteria of how knowledge should be selected for rehearsal over training. These critical aspects collectively contribute to a comprehensive understanding of the \ac{CPD} algorithm's efficacy in the context of continual learning for in-hand manipulation tasks.

    \section{Related Works}
        \subsection{Learning Control Policies for Soft Robots}
            Learning-based methods encompass the empirical approximation of an unknown model of the soft robot or control policies \cite{laschi2023learningcontrol} by relying on supervised \cite{Nazeer2023} or \ac{RL} techniques \cite{centurelli2022dynamic,Bianchi2023}. Model-free controllers \cite{donato2022plantmovements,donato2023plantreaching} have also been proposed to address the redundancy issue. \ac{RL} facilitates the learning of kinematic/dynamic models and enables the direct acquisition of the controller itself \cite{thuruthel2019rl}. However, solutions like online \ac{RL} can be resource-intensive, requiring a significant number of interactions with the environment and extensive computational resources \cite{yu2022manipulationRL}. 


        \subsection{Policy Distillation}
            Knowledge distillation is a model compression technique that transfers knowledge from a larger, complex model (teacher) to a smaller, simpler model (student) \cite{hinton2015distillation}. The student model is trained to learn the decision-making process of the teacher model by matching a set of output activations or feature maps produced by the teacher. \ac{PD} is a specific application of knowledge distillation that aims to extract the policy of an \ac{RL} agent \cite{rusu2016policy}. \ac{PD} considers not only the expert's actions but also the teacher's policy distribution, allowing for a more comprehensive transfer of knowledge by learning from the teacher's exploration and exploitation strategies. \ac{PD} has already been applied to robotic \ac{RL}-based controllers \cite{jayanthi2023distillation}, addressing how to deal with heterogeneity and generalizability during distillation.

        \subsection{Continual Learning in Robotics}
            \ac{CL} addresses scenarios where the data distribution and learning objective change over time or where the entire training data and objective criteria are not available simultaneously \cite{lesort2020clrobotics}. Among the many challenges addressed by \ac{CL}, there are: \textit{catastrophic forgetting}, when previously learned knowledge is forgotten over incremental fine-tuning of the target model; \textit{memory handling} to store and retain important information from past tasks; \textit{detection of distributional drifts} due to changes in the input distribution or the introduction of new classes or tasks. Robots are an ideal domain for \ac{CL} application: robots cannot revisit past experiences to improve their prior learning, emphasizing the need for continual improvement \cite{pique2022clsoft}. Moreover, \ac{CL} is well-suited to optimize the power and memory constraints often faced by robots, thereby enhancing their learning capabilities. Traoré et al. \cite{traore2019discorl} propose an approach similar to our \ac{CPD} framework, but they learn over discrete action and observation spaces and assume an unlimited replay buffer.
    
    \section{Methodology}
\begin{figure}[b!]
            \includegraphics[width=\linewidth]{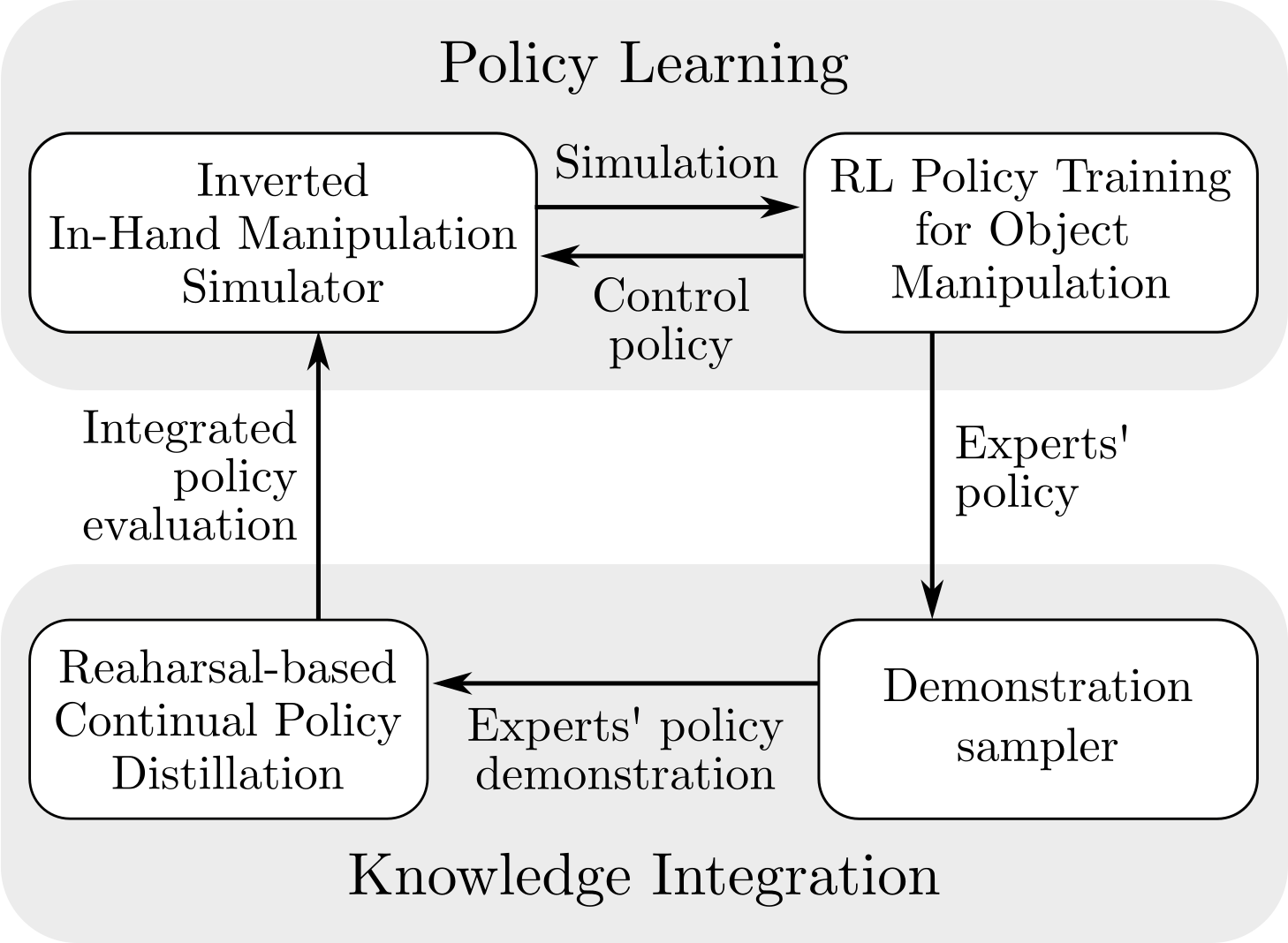}
            \captionsetup{justification=centering}
            \caption{ \ac{CPD} framework pipeline}
            \label{fig:workflow}
        \end{figure}

        The proposed \ac{CPD} framework is established through a pipeline shown in Fig.\ref{fig:workflow}, that involves recurrently \textit{learning} a soft robotic hand controller in simulation for in-hand manipulation of a specific object, and its \textit{integration} with the previously acquired knowledge to build a versatile controller. Importantly, this integration process must be accomplished without access to the pre-training data, ensuring data privacy.
        
        During Policy Learning, the expert's control policy undergoes training using \ac{RL} for in-hand re-orientation of an object. This process can be repeated for various objects, with a distinct expert trained for each. In Knowledge Integration, demonstrations of the experts' policies are asynchronously sampled and employed for rehearsal-based continual policy distillation. This entails integrating the acquired knowledge from multiple experts into a unified policy. The policy evaluation is conducted within the simulated environment. A glossary of the terminology is available in Appendix \ref{appendix:glossary}.
        
        \subsection{In-Hand Manipulation Task}
            The \ac{IHM} task is performed using SoMo \cite{graule2021somo}, an open-source framework designed for simulating soft/rigid hybrid systems within customizable \ac{RL} training environments through SoMoGym \cite{graule2022somogym}. SoMo approximates continuum manipulators using rigid-link systems with spring-loaded joints and it is implemented on PyBullet \cite{coumans2020pybullet}, handling also inter-body contacts.

            \ac{IHM} is a significant challenge for every dexterous robot, and the soft domain makes it even harder to deal with from both a design and control point of view. A four-fingered soft gripper is integrated into SoMo, featuring two independently controlled DoFs for each finger. The fingers are linked by a rigid palm, which aids in achieving precise manipulation within the gravitational field while orienting the gripper in an upward direction. As it can be observed in Fig. \ref{fig:simulators}, flexion/extension enables movement of the finger towards or away from the palm, while adduction/abduction allows for left or right lateral motion. 

\begin{figure}[t!]
                \centerline {\includegraphics[width=\linewidth]{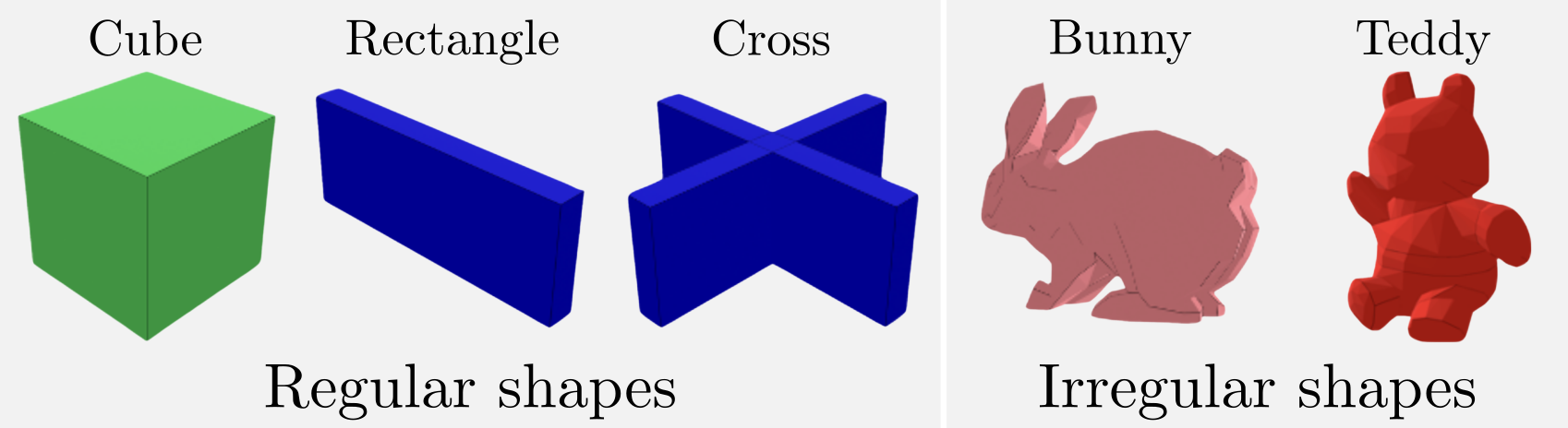}}
                \captionsetup{justification=centering}
                \caption{Objects used for In-Hand Manipulation}
                \label{fig:objects}
            \end{figure}

            A variety of objects have been introduced in the \ac{IHM} environment to allow for a comprehensive evaluation of the framework, from regular to irregular shapes as in Fig.\ref{fig:objects}. All the objects are not deformable.
            
        \subsection{Learning Control Policy via Reinforcement Learning} \label{sec:method-rl}
            From the pool of \ac{RL} algorithms implemented in Stable-Baselines3 \cite{raffin2021stablebaselines}, the \ac{PPO} algorithm is used to learn the policy for the \ac{IHM} task in our custom SoMoGym environment. At each time-step $t$, the object is characterised by a position $[x_t, y_t, z_t] \in \Re^3$ and orientation $[\phi_t, \psi_t, \theta_t] \in \Re^3$, where $\phi$, $\psi$, $\theta$ are the angles of rotation around respectively $z$, $y$, and $x$ axes. The \ac{PPO}'s reward function accounts for the variations of the object's pose over time, and highly promotes counter-clockwise rotations around the orthogonal axis to the palm, yet maintaining a secure grip to prevent the object from dropping.
            \begin{equation*}
                \begin{split}
                    R_t = & 1000 (\phi_t-\phi_{t-1}) - |\psi_t-\psi_{t-1}| - |\theta_t-\theta_{t-1}|\\
                    & - |x_t - x_{t-1}| - |y_t - y_{t-1}|
                \end{split}
            \end{equation*}
            We determined the constants for the reward function through empirical evaluations, but they can be further adjusted using a grid search. In contrast to the reward function in \cite{graule2022somogym}, we loosened the restrictions on positional changes, penalizing only changes along the object's $x$ and $y$ axes, and focused on enhancing the orientation at each step rather than considering cumulative orientation changes. During each training episode, the episodic reward and relevant components such as the object's pose are recorded.
            
            To evaluate each control policy, new environments are initialized with same objects for manipulation and same initial pose but unique random seeds to examine the generalization capabilities of the controllers. The \ac{PPO} agent with the highest score is saved, along with checkpoint models at predefined intervals. To select the best object-specific manipulation policies as the experts for knowledge integration, we consider the following criteria: (i) highest evaluation episodic reward; (ii) related stable training episodic reward history; (iii) demonstration of a complex \ac{IHM} gait, where the fingers repeatedly establish and release contact with the manipulated object; (iv) demonstration of a unique gait.
        
        \subsection{Offline Experts' Demonstrations and Policy Distillation}            
            Imitation Learning focuses on training a policy network to replicate the expert's demonstrated behaviour by optimizing the alignment of observed states with corresponding expert actions. Conversely, \ac{PD} involves using the expert policy to generate state-action pairs, which are then utilized to train a more efficient student policy network that learns the decision-making process of the expert policy. Our goal is to merge the policies of multiple experts into a single distilled agent that captures their common behaviours. Therefore, we combine the two concepts by using supervised learning and the demonstrations provided by experts to develop a robust and adaptive policy.
            
            To evaluate the offline \ac{PD} task, we consider three loss functions $\ell$ for matching action distributions of experts \cite{rusu2016policy}. 
            \begin{itemize}
                \item Mean Squared Error (MSE)
                    \begin{equation*}
                        {\ell}_{mse}(\pi,{s}^{\star},{a}^{\star})={\lVert(\pi({s}^{\star})-{a}^{\star})\rVert}^2_2
                    \end{equation*}
                \item Negative Log-Likelihood (NLL)
                    \begin{equation*}
                        {\ell}_{nll}(\pi,{s}^{\star},{a}^{\star})=-\ln\pi({a}^{\star}\arrowvert{s}^{\star})
                    \end{equation*}
                \item Kullback-Leible Divergence (KL)
                    \begin{equation*}
                        {\ell}_{kl}(\pi,{\pi}^{\star})=\log{\frac{{\sigma}^{\star}}{\sigma}}+\frac{{\sigma}^2+{(\mu-{\mu}^{\star})}^2}{2{{\sigma}^{\star}}^{2}}-\frac{1}{2}
                    \end{equation*}
            \end{itemize}
            $a$ and $s$ denote the actions and observation states generated from our target distilled policy $\pi$, while $\mu$ and $\sigma$ represent the mean and standard deviation of our target distilled policy. We denote with $\star$ all the entities related to the experts' policies.
            
            For each i-th expert, we utilize their control policies to sample demonstrations ${\mathcal{D}}_{{\pi}^{\star}_i}$ following the distribution $d^{{\pi}^{\star}_i}$ with length $M_i$:    
            \begin{equation}
                {\mathcal{D}}_{{\pi}^{\star}_i} = {({s}^{\star}_j, {a}^{\star}_j)}^{M_i}_{j=1} \sim d^{{\pi}^{\star}_i}
                \label{eq:dem}
            \end{equation}
            Expert policies may not always be accessible, and each expert policy has its own $\mu_i$ and $\sigma_i$. In the case of ${\ell}_{kl}$ as loss function, an approximation method is employed: each sampled action $a^{\star}_j$ generated by expert policy ${\pi}^{\star}_i$ is considered as the mean of ${\pi}^{\star}_i$'s action distribution with a fixed small standard deviation ${\sigma}^{\star}={1e}^{-6}$, representing a deterministic action distribution.
            
        \subsection{Knowledge Integration: Continual Policy Distillation} \label{sec:cpd}
            When generating the experts' demonstrations, we assume no limitation on when we can obtain the sampled demonstrations or how many we can store. However, in a real-case scenario, we are constrained by the size of the memory buffer $M$ used to store the generated demonstrations and when the experts would be available. Considering the chronological order, we may need to delete old acquisitions to make space for new ones. Given these considerations, we make the following assumptions before initiating the \ac{PD} process: (i) object-specific environments are accessible only to evaluate the control policy; (ii) experts are available only for generating demonstrations; (iii) experts' demonstrations are generated in chronological order and sequentially accessible; (iv) the storing memory size $M$ for demonstrations is limited. Based on the aforementioned assumptions, the \ac{PD} process aligns with the \ac{CL} paradigm, and we leverage the Avalanche framework \cite{lomonaco2021avalanche} to facilitate its implementation.
            
            Within the \ac{CL} paradigm, the experts' demonstrations can be seen as a continual stream of data composed of a series of experiences, each representing an expert demonstration for manipulating a specific object. Based on the division of incremental batches and the availability of task IDs (e.g. object recognition), we operate within the \textit{domain-incremental learning} scenario, as the experiences share the same action space but have different distributions \cite{vandeven2019clscenarios}. Task IDs are now deducted from the observation space, but real-world applications may rely on sensory information.

            To design the \ac{CPD} process, the first step is defining the target student control policy. Instead of compressing expert policies, the approach focuses on integrating them. This is achieved by creating a blank student policy with the same architecture as the expert policy, treating it as a \ac{PPO} agent. Distilling the expert policies directly into the blank student policy is done through \ac{PD} using supervised learning. The expert observation is the input, and the expert action is the output. After training, the distilled policy network is reinserted into the student agent for evaluation.
            
            The model's performance may deteriorate on previously learned concepts as it sequentially learns new concepts. To tackle this problem, various \ac{CL} strategies \cite{lesort2020clrobotics} (e.g., weight regularization, model capacity expansion, synthetic data generation) have been proposed to mitigate or prevent catastrophic forgetting. In this work, we focus on \textit{rehearsal-based methods} \cite{chaudhry2019reheasalcl}, as they effectively improve model generalization. These methods involve storing old data and replaying it during the training process with new data. In this way, it can retain its performance on previous tasks while learning new ones. Specifically, our method is based on exemplar-based rehearsal methods rather than generative-based ones. Exemplar-based rehearsal methods store a subset of previously seen data, known as \textit{exemplars}, and utilize them to train the model. The goal is to store the most informative examples from each observed manipulation task and ensure that the exemplars are diverse enough to cover the underlying distribution of the previous data.

            \begin{algorithm}
                \caption{Policy Distillation with Exemplar-based Rehearsal Methods}
                \label{alg:CPD}
                \begin{algorithmic}[1]
                    \REQUIRE
                    Experience buffer $B_{\text{exp}}$  with size $M$
                    \ENSURE
                    Distilled target policy $\hat{\pi}$
                    \WHILE{a new experience exists in the experience sequence}
                    \STATE Add the demonstration of the current experience obtained through Eq.\ref{eq:dem} to the experience buffer $B_{\text{exp}}$;
                    \STATE Subsample each expert demonstration ${\mathcal{D}}_{{\pi}^{\star}_i}$ stored in the experience buffer $B_{\text{exp}}$  with size $M_i$ using the selected replay strategy, under the condition that $M=\sum_{i=1}^N M_i$;
                    \STATE Perform offline supervised learning from $B_{\text{exp}}$  by minimizing the imitation loss according to the chosen criterion: $\hat{\pi}=\min_{\pi}\sum_{i=1}^N\sum_{j=1}^{M_i}\ell(\pi(s_j^{\star},a_j^{\star}))$;
                    \ENDWHILE
                    \RETURN{Distilled target policy $\hat{\pi}$}
                \end{algorithmic}
            \end{algorithm}
                
            Building upon the previous discussions, we propose a formalization of the \ac{CPD} algorithm in Alg.\ref{alg:CPD} by employing the \ac{PD} approach along with exemplar-based rehearsal methods. The selection of replay strategies for maintaining the experience buffer $B_{\text{exp}}$ determines which expert demonstrations are used for \ac{CPD}. We differentiate the replay strategies on their storage management, which involves organizing the most representative demonstrations as a core set within the limited size of the replay buffer.
            
            \begin{itemize}
                \item \textit{Experience Balanced Replay (ReplayBR)}. It handles a rehearsal buffer 
                \begin{equation*}
                    B_{\text{exp}} = \sum_{i=1}^{N} \mathcal{D}_{{\pi}^{\star}_i} 
                \end{equation*}
                with size $M$ and achieves a balanced selection of samples across experiences, ensuring that $M_i = |\mathcal{D}_{{\pi}^{\star}_i}| = M/N$, where $M_i$ denotes the size of demonstrations for i-th experience. $M$ represents the total size of the experience replay buffer, and $N$ is the number of encountered experiences.
                \item \textit{Ex-Model Experience Replay (ReplayEX)} \cite{carta2021exmodel}. This storage strategy differs from ReplayBR by randomly subsampling demonstrations from the current buffer and the new experience. The buffer is defined as
                \begin{equation*}
                    B_{\text{exp}_i} = \text{subsample}(B_{\text{exp}_{i-1}}) + \text{subsample}(\mathcal{D}_{{\pi}^{\star}_i})
                \end{equation*}
                Here, $|B_{\text{exp}_i}| = |B_{\text{exp}_{i-1}}| = M$ represents the size of the replay buffer upon the arrival of the $(i-1)$-th and $i$-th experiences, respectively. $|\mathcal{D}_{{\pi}^{\star}_i}| = M_i$ represents the size of the $i$-th experience.
                \begin{equation*}
                    |\text{subsample}(B_{\text{exp}_{i-1}})| = M - M/M_i
                \end{equation*}
                denotes the size of the subsampled replay buffer at the $(i-1)$-th experience, and 
                \begin{equation*}
                    |\text{subsample}(\mathcal{D}_{{\pi}^{\star}_i})| = M/M_i
                \end{equation*}    
                represents the new subsampled $i$-th experience size. 
                \item \textit{Reward Prioritized Experience Replay (ReplayRP)}. As a greedy ReplayBR strategy, it aims to maximize the episodic rewards by selecting exemplars in descending order of their episodic rewards. This approach gives priority to storing higher rewarding exemplars rather than focusing on maintaining a diverse set in the limited memory buffer. Such strategy relies on having access to episodic rewards along with demonstrations.
                \item \textit{Reward Weighted Reservoir Sampling Experience Replay (ReplayRPR)}. An additional step is introduced to ReplayRP to ensure exemplar selection diversity to cover the underlying distribution. This is achieved by incorporating randomness into the greedy subsampling method through reward-weighted reservoir sampling based on episodic rewards. 
            \end{itemize}
            
            To provide context for the results of the proposed \ac{CL} strategies, we established few baseline strategies as references for comparison.
            \begin{itemize}
                \item \textit{Naive Task Incremental Learning}. It fine-tunes a single model incrementally without any specific method to address catastrophic forgetting of previous knowledge.
                \item \textit{Joint Training}. It performs offline training by accessing all experts' demonstrations at the same time. Joint Training is implemented by Cumulative Training: the model is trained with the accumulated data from all previously encountered experiences and the current experience. The average task score up to the current experience provides an aggregated measure of performance across all tasks encountered thus far, including the current experience.
            \end{itemize}
    
    \section{Experimental Results}    
        \subsection{Expert Control Policy Learning}
            To expedite the training process, we concurrently execute five separate instances of the Inverted \ac{IHM} environment: we leverage the \ac{PPO}'s ability to run multiple workers simultaneously to decrease the overall training time. Each environment undergoes $4 \times 10^6$ steps. The duration of a single run is dependent on the available computing resources, typically from 1.5 to 4.5 days and requiring approximately 13GB of RAM and 1GB of graphic memory. The final trained controller and the one with the best evaluation score are retained. To ensure both training repeatability and generalization, we conduct twenty seed-changing runs for each object. 

\begin{figure}[b!]
                \centerline {\includegraphics[width=\linewidth]{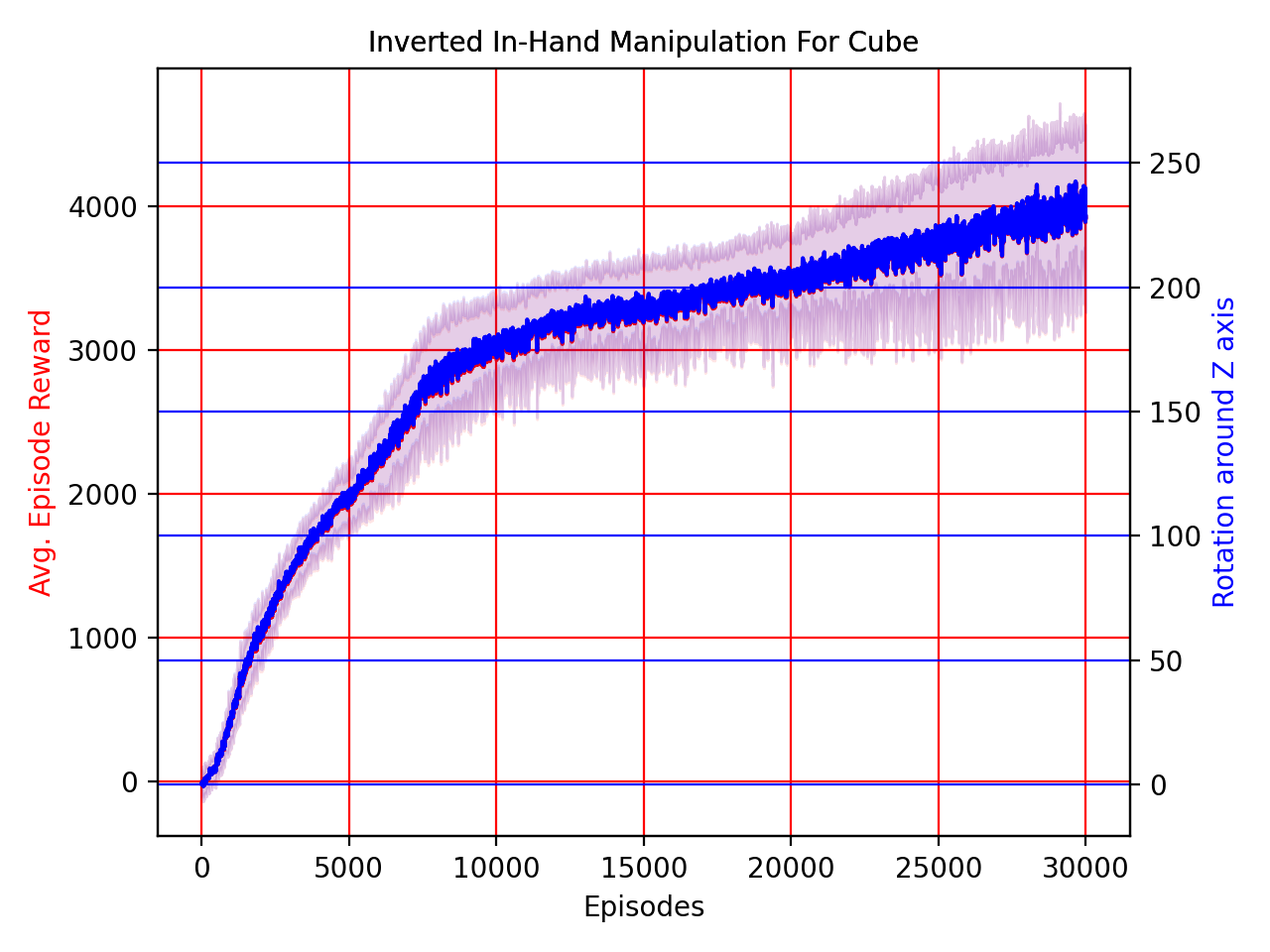}}
                \captionsetup{justification=centering}
                \caption{Learning curve for cube manipulation}
                \label{fig:rl_cube_res}
            \end{figure}

            As an illustrative example, we focus on the manipulation of the cubic shape. Fig. \ref{fig:rl_cube_res} shows the learning curves recorded across five different random seeds throughout the training phase and representing the average smoothed episode reward and rotation around the z-axis in degrees. The discrepancy between the learning curves is negligible, supporting the use of rotation around the z-axis as a reliable metric for evaluating the performance of the controllers.
            
            \begin{figure}[b!]
                \centerline {\includegraphics[width=\linewidth]{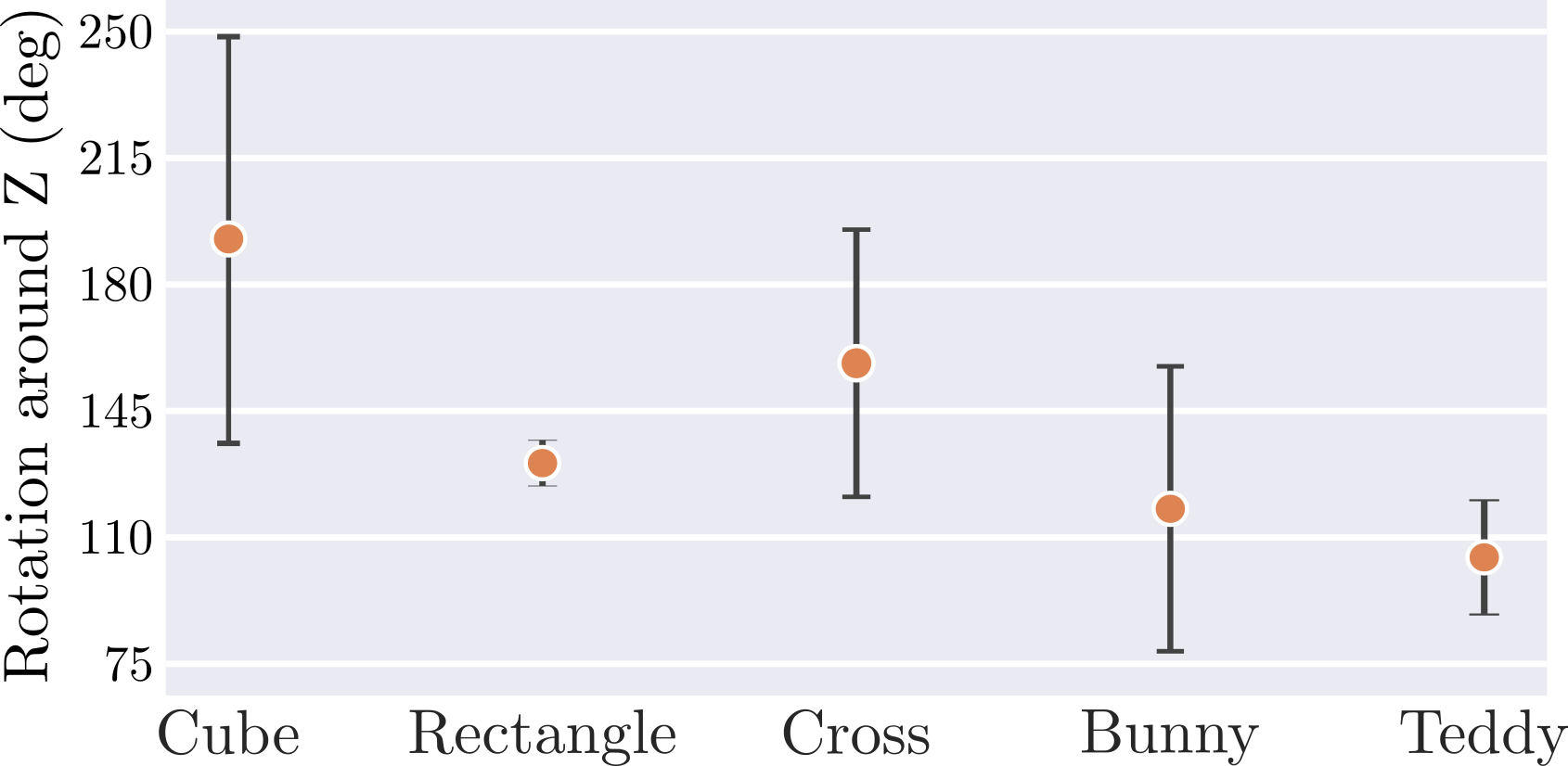}}
                \captionsetup{justification=centering}
                \caption{Average performance of experts' policy}
                \label{fig:rotations}
            \end{figure}

            \begin{figure}[b!]
                \centerline {\includegraphics[width=\linewidth]{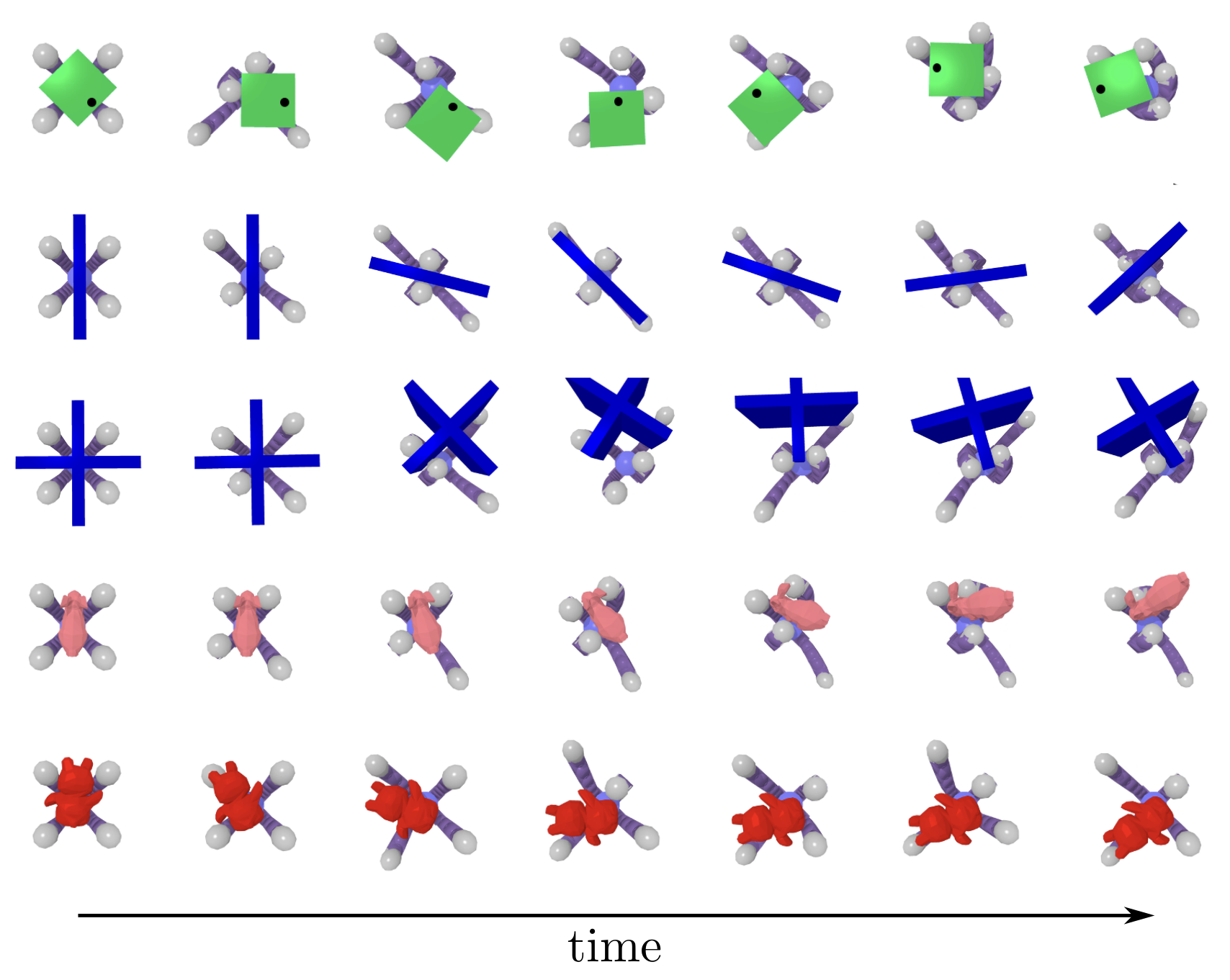}}
                \captionsetup{justification=centering}
                \caption{Snapshots of the experts' policy rollout over one episode for each object}
                \label{fig:rollout}
            \end{figure}

            The selection of expert controllers adheres to the criteria outlined in Sec. \ref{sec:method-rl}. Results from multiple runs, with different random seeds, are averaged and presented in Fig. \ref{fig:rotations}. Independent manipulation tasks exhibit varying levels of complexity, resulting in diverse task scores and highlighting the challenges associated with each object. It emphasizes the need for tailored approaches to address individual tasks.

            Looking at Fig.\ref{fig:rollout}, the robot employs grasping and re-grasping actions to handle different objects. Distinct rotation methods are tailored to the objects' geometry: for the rectangular shape, one pair of fingers maximally extends for smooth top-down object passage, while the other assists in rotation. For cross-shaped objects, instead, they employ a multi-step approach, pushing the object to the palm's edge to introduce gravitational interference and applying a sequence of grasps and re-grasps on the object's edge to enable rotation. Dealing with objects that have irregular shapes is more challenging because of stability. For instance, with the bunny, it extends one finger to clear the object from above and relies on the others for rotation. When handling the teddy bear, it first leans the object against its fingers and then moves the fingers underneath to make it rotate. 

            These characteristics align with the criteria defined in Sec. \ref{sec:cpd} and make the expert policies suitable for the \ac{CPD}. Sequentially learning such behaviours helps in observing the occurrence of catastrophic forgetting. However, it is worth noting that the control policies for irregular objects introduce additional uncertainty due to potential loss of contact, which can render it less stable compared to that for manipulating regular objects.
        
        \subsection{Continual Policy Distillation}
            We generate $10^3$ episodes of demonstrations for each expert policy as explained in Eq. \ref{eq:dem} to create the \ac{CPD} input dataset. To examine the effectiveness of different \ac{PD} loss functions, we design experiments with consistent training parameters following a coarse-grained grid search. Tab. \ref{tab:loss} presents the performance of the Joint Training baseline using three different loss functions. Training with the KL and MSE learning criteria outperforms training with the NLL criterion. Specifically, the KL achieves the highest performance, aligning with its capability to capture the crucial decision-making regions of the target policy and effectively reproduce its behaviour. As a result, the KL will be utilized as loss function for the offline \ac{CPD}.

            We employ the Naive Incremental Learning (lower bound) and the Cumulative Training (upper bound) baselines for performance evaluation, whose performance is reported in Tab. \ref{tab:baselines}. For Naive Incremental Learning, the average task score across all tasks encountered and the current experience serve as an illustration of the catastrophic forgetting phenomenon. As the experiences progress, the averaged task score gradually decreases, indicating that the agent's performance in previously learned tasks diminishes as it learns new tasks. Despite local optimal performance, it highlights the challenge of retaining knowledge from previous experiences while accommodating new learning.

            We evaluate the effectiveness of different \ac{CL} strategies across various settings, by conducting a comprehensive analysis of the performance of the Alg. \ref{alg:CPD}. We systematically vary the size M of the replay buffer $B_{\text{exp}}$ to: M=$10^3$, the size of a single expert demonstration; M=$10^2$, or $10\%$ of the size of a single expert demonstration; M=$10^1$, or $1\%$ of the size of a single expert demonstration. Tab. \ref{tab:CPD_M} showcases \ac{CL} performance, and Naive and Cumulative baselines are reported for ease of comparison. Likewise, each row represents the average task score across all encountered experiences and the current experience. The last row of each section provides the averaged performance for each \ac{CL} strategy by across all experiences. All task scores are averaged across ten runs with different random seeds, ensuring robustness in the evaluation.

            \begin{table}[t!]
                \centering
                \resizebox{\linewidth}{!}{\begin{tabularx}{1.15\linewidth}{|X|X|X|X|X|X|X|}
                    \hline \centering
                    \multirow{2}{*}{Loss} & \multicolumn{6}{c|}{Control policy performance [avg$\pm$std] [degrees]} \\ 
                    \cline{2-7} & \multicolumn{1}{c|}{Cube} & \multicolumn{1}{c|}{Rect} & \multicolumn{1}{c|}{Bunny} & \multicolumn{1}{c|}{Cross} & \multicolumn{1}{c|}{Teddy} & \multicolumn{1}{c|}{Avg} \\ 
                    \hline
                    \centering MSE & 214$\pm$51 & \bf 128$\pm$5 & 115$\pm$21 & \bf 129$\pm$14 & 113$\pm$20 & 140\\
                    \hline
                    \centering NLL & 100$\pm$44 & 125$\pm$4 & 100$\pm$18 & 93$\pm$35 & 107$\pm$24 & 105\\
                    \hline
                    \centering KL & \bf 224$\pm$40 & 127$\pm$5 & \bf 122$\pm$8 & 120$\pm$11 & \bf 118$\pm$22 & \bf 142\\
                    \hline
                \end{tabularx}}
                \caption{Cumulative Training using different loss functions}
                \label{tab:loss}
            \end{table}

            \begin{table}[t!]
                \centering
                \resizebox{\linewidth}{!}{\begin{tabularx}{1.15\linewidth}{|X|X|X|X|X|X|X|}
                    \hline \centering
                    \multirow{2}{*}{Exp} & \multicolumn{6}{c|}{Task performance [avg$\pm$std] [degrees]} \\ 
                    \cline{2-7} & \multicolumn{1}{c|}{Cube} & \multicolumn{1}{c|}{Rect} & \multicolumn{1}{c|}{Bunny} & \multicolumn{1}{c|}{Cross} & \multicolumn{1}{c|}{Teddy} & \multicolumn{1}{c|}{Avg} \\ \hline
                    \hline \multicolumn{7}{|c|}{\textbf{Naive Incremental Learning}}\\
                    \hline \centering
                    Cube & \bf 242$\pm$29 & 48$\pm$24 & 20$\pm$7 & 42$\pm$9 & 55$\pm$43 & \bf 242 \\
                    \hline \centering
                    Rect & 56$\pm$1 & \bf 128$\pm$4 & 93$\pm$14 & 26$\pm$1 & -5$\pm$1 & 92 \\
                    \hline \centering
                    Bunny & 39$\pm$3 & 40$\pm$8 & \bf 108$\pm$22 & 24$\pm$4 & -2$\pm$0 & 62 \\
                    \hline \centering
                    Cross & 0$\pm$8 & -32$\pm$53 & 16$\pm$10 & \bf 128$\pm$7 & 47$\pm$44 & 28 \\
                    \hline \centering
                    Teddy & -4$\pm$5 & -14$\pm$21 & 2$\pm$2 & -5$\pm$12 & \bf 126$\pm$9 & 21 \\\hline
                    \hline \multicolumn{7}{|c|}{\textbf{Cumulative Training}}\\
                    \hline \centering
                    Cube & \bf 243$\pm$24 & 32$\pm$20 & 33$\pm$25 & 45$\pm$19 & 52$\pm$42 & \bf 243\\
                    \hline \centering
                    Rect & \bf 218$\pm$40 & \bf 129$\pm$4 & 47$\pm$35 & 39$\pm$12 & 50$\pm$54 & \bf 173\\
                    \hline \centering
                    Bunny & \bf 227$\pm$54 & \bf 126$\pm$5 & \bf 110$\pm$19 & 25$\pm$5 & 8$\pm$33 & \bf 154\\
                    \hline \centering
                    Cross & \bf 213$\pm$40 & \bf 129$\pm$1 & \bf 105$\pm$16 & \bf 126$\pm$47 & 2$\pm$50 & \bf 143\\
                    \hline \centering
                    Teddy & \bf 205$\pm$40 & \bf 129$\pm$2 & \bf 120$\pm$23 & \bf 122$\pm$22 & \bf 117$\pm$10 & \bf 139\\
                    \hline
                \end{tabularx}}
                \caption{Performance baselines}
                \label{tab:baselines}
            \end{table}

            \begin{table}[t!]
                \centering
                \resizebox{\linewidth}{!}{\begin{tabularx}{1.15\linewidth}{|X|X|X|X|X|X|X|}
                    \hline \centering
                    \multirow{2}{*}{Exp} & \multicolumn{6}{c|}{Knowledge integration performance [avg] [degrees]} \\ 
                    \cline{2-7} & \multicolumn{1}{c|}{Naive} & \multicolumn{1}{c|}{ReplayBR} & \multicolumn{1}{c|}{ReplayRP} & \multicolumn{1}{c|}{ReplayRPR} & \multicolumn{1}{c|}{ReplayEX} & \multicolumn{1}{c|}{Cumul}\\ \hline \hline \multicolumn{7}{|c|}{\textbf{Replay Buffer Size M = $\mathbf{10^3}$}}\\
                    \hline
                    Cube & 242 & 191 & \bf 230 & 191 & 215 & \bf 243 \\
                    \hline
                    Rect & 92 & 188 & \bf 190 & 167 & 182 & 173 \\
                    \hline
                    Bunny & 62 & 152 & 148 & \bf 154 & 149 & \bf 154 \\
                    \hline
                    Cross & 28 & 140 & \bf 155 & 151 & 143 & 143 \\
                    \hline
                    Teddy & 21 & 137 & 146 & 137 & \bf 149 & 139 \\
                    \hline
                    Avg & 89 & 162 & \bf 174 & 160 & 168 & 170 \\\hline \hline \multicolumn{7}{|c|}{\textbf{Replay Buffer Size M = $\mathbf{10^2}$}}\\         \hline
                    Cube & 242 & 214 & \bf 223 & 214 & 217 & \bf 243 \\
                    \hline
                    Rect & 92 & 163 & \bf 180 & 179 & 169 & 173 \\
                    \hline
                    Bunny & 62 & 151 & 158 & \bf 159 & 156 & 154 \\
                    \hline
                    Cross & 28 & 145 & 142 & \bf 154 & 128 & 143 \\
                    \hline
                    Teddy & 21 & 132 & 136 & \bf 144 & 110 & 139 \\
                    \hline
                    Avg & 89 & 161 & 168 & \bf 170 & 156 & \bf 170 \\
                    \hline \hline \multicolumn{7}{|c|}{\textbf{Replay Buffer Size M = $\mathbf{10^1}$}}\\
                    \hline
                    Cube & 242 & \bf 240 & 216 & \bf 240 & 96 & \bf 243 \\
                    \hline
                    Rect & 92 & \bf 182 & 177 & 156 & 65 & 173 \\
                    \hline
                    Bunny & 62 & 129 & \bf 145 & 117 & 95 & \bf 154 \\
                    \hline
                    Cross & 28 & 111 & \bf 119 & 117 & 78 & \bf 143 \\
                    \hline
                    Teddy & 21 & 92 & \bf 110 & 85 & 72 & \bf 139 \\
                    \hline
                    Avg & 89 & 151 & \bf 153 & 143 & 81 & \bf 170 \\
                    \hline
                \end{tabularx}}
                \caption{Comparison of replay-based \ac{CL} strategies with variable size replay buffer}
                \label{tab:CPD_M}
            \end{table}

            ReplayRP exhibits comparable performance to the upper bound cumulative strategy; this finding is expected since ReplayRP aims to maintain the replay buffer with data generated from episodes with the highest episodic rewards. The ReplayEX demonstrates favourable performance when a relatively large buffer size is available: this is attributed to ReplayEX's ability to incorporate a diverse set of previously encountered data and current data, leveraging its storage strategy effectively. However, as the replay buffer size decreases, ReplayEX is unable to fully capitalize on the data diversity, leading to a decline in performance. ReplayRPR achieves its best performance when the buffer size is M=$10^2$. In this scenario, ReplayRPR outperforms ReplayRP by incorporating randomness during reward-prioritized sampling. This introduces an element of exploration, allowing for the selection of diverse experiences to improve the performance. ReplayBR, instead, demonstrates comparable performance when a sufficiently large replay buffer size is provided (preferably larger than $10\%$ of the length of a single experience). Notably, the performance of all replay strategies deteriorates as the replay buffer size decreases.
    
    \section{Discussion}
        We aimed to develop a versatile soft robotic hand controller capable of manipulating distinct objects efficiently and effectively. The proposed \ac{CPD} framework involves training from multiple expert controllers sequentially for specific object rotation tasks. Instead of accessing the control agent directly, we utilized policy rollouts as expert demonstrations, which were stored in a limited-size memory buffer.

        Fig. \ref{fig:rotations} and Fig. \ref{fig:rollout} showcase how object shape influences the controller's performance. Objects such as rectangles and cross shapes exhibit significantly different dimensions compared to the cube object, with lengths approximately 2.5 times longer in certain dimensions. This size difference can impact the controller's behaviour, potentially leading it to prioritize strategies that involve extending the fingers to accommodate the longer dimensions for smoother object passage. Furthermore, irregular shapes like the bunny and teddy introduce additional complexities not encountered with basic shapes like cubes and rectangles. The irregular contours of these objects can challenge the soft robotic hand in establishing symmetrical torques necessary for a stable grip or maintaining control during manipulation. Unlike regular shapes where geometric symmetry aids manipulation, irregular shapes may require the controller to adapt its strategy to accommodate asymmetrical features, making learning the manipulation strategy more challenging to generalize. 
        
        In this context, utilizing object pose as the primary reward feedback provides a straightforward and interpretable metric for evaluating the policy's effectiveness. However, the variations in object shapes emphasize the significance of considering shape characteristics to attain robust and versatile manipulation capabilities. In real-world scenarios where vision is constrained or unreliable, incorporating additional localized sensory feedback, such as tactile sensing, can assist the controller in overcoming these limitations and executing effectively.
    
        The experimental results shed light on the critical significance of choosing an appropriate replay strategy, particularly tailored to the size of the available buffer and the trade-off between exploration and exploitation. Understanding the strengths and limitations of each strategy is the key to effectively address the catastrophic forgetting and develop a versatile controller. Our contribution also underscores the importance of selecting suitable loss functions for distillation within the \ac{CPD} framework. 

        In the context of time efficiency, replay-based \ac{CPD} approaches provide a substantial advantage over online \ac{RL}. Online \ac{RL} typically demands a longer time to train from scratch, making it a more time-intensive process, instead of learning from expert demonstrations like in the \ac{CPD} framework. Efficiency represents a clear distinction between the two approaches and underscores the practical benefits of employing \ac{CPD} in the development of soft robot controllers.

        The \ac{CPD} approach does not require any data used to train the experts. Instead, it exploits experts to generate input demonstrations, making this approach advantageous in terms of respecting the privacy of training data. However, it still relies on the availability of an environment to generate demonstrations. Further implementations will address these limitations by building a generative model while training the \ac{RL}-based expert controller, rather than relying on direct interaction with the environment.
        
        In addition, embedding the object \ac{ID} into the observation space is a brute-force method of providing information to the agent for recognizing different tasks. The presence of object \ac{ID} implies knowledge of future tasks in advance, which limits the applicability to unknown objects. Sensory-based recognition will allow the automatic recognition of further objects without any prior knowledge.

    \section{Conclusion}
        This work contributes to the advancement of soft robotic manipulation by integrating \ac{RL} and \ac{CL}. By addressing the catastrophic forgetting problem and integrating knowledge from multiple \ac{RL} agents, the developed soft robotic hand controller enables versatile and adaptive behaviours, facilitating the effective deployment of soft robotic systems in various scenarios.
        
        The approach offers advantages in terms of time efficiency, with significantly reduced training time compared to traditional online \ac{RL}. By utilizing a limited-size memory buffer, memory efficiency is optimized while retaining important knowledge for learning. 

        Overall, \ac{CPD} provides a practical and efficient sequential learning framework to address the challenge of knowledge integration in complex manipulation tasks. It contributes to the advancement of learning-based control strategies for soft robotic systems within the paradigm of \ac{CL}. \ac{CPD} holds promise for developing intelligent and adaptive soft robotic hand controllers that continually acquire and retain knowledge, enabling them to perform a wide range of tasks effectively in several applications.
        
    \printbibliography

@inproceedings{graule2021somo,
  title={SoMo: Fast and Accurate Simulations of Continuum Robots in Complex Environments},
  author={Graule, M.A. and Teeple, C.B. and McCarthy, T.P. and Kim, G.R. and St. Louis, R.C. and Wood, R.J.},
  booktitle={IEEE-RSJ IROS},
  year={2021},
  doi={10.1109/IROS51168.2021.9636059}
}

@article{graule2022somogym,
  title={SoMoGym: A Toolkit for Developing and Evaluating Controllers and Reinforcement Learning Algorithms for Soft Robots},
  author={Graule, M.A. and McCarthy, T.P. and Teeple, C.B. and Werfel, J. and Wood, R.J.},
  journal={IEEE Robotics and Automation Letters},
  volume={7},
  number={2},
  pages={},
  year={2022},
  doi = {10.1109/LRA.2022.3149580}
}

@inproceedings{lomonaco2021avalanche,
  title={Avalanche: An End-to-End Library for Continual Learning},
  author={Lomonaco, V. and Pellegrini, L. and Cossu, A. and Carta, A. and Graffieti, G. and Hayes, T.L. and De Lange, M. and Masana, M. and Pomponi, J. and {van de Ven}, G.M. and Mundt, M. and She, Qi and Cooper, K. and Forest, J. and Belouadah, E. and Calderara, S. and Parisi, G.I. and Cuzzolin, F. and Tolias, A.S. and Scardapane, S. and Antiga, L. and Ahmad, S. and Popescu, A. and Kanan, C. and {van de Weijer}, J. and Tuytelaars, T. and Bacciu, D. and Maltoni, D.},
  booktitle={IEEE Conference on Computer Vision and Pattern Recognition Workshops (CVPRW)},
  year={2021},
  doi={10.1109/CVPRW53098.2021.00399}
}

@inproceedings{carta2021exmodel,
  title={Ex-Model: Continual Learning from a Stream of Trained Models},
  author={Carta, A. and Cossu, A. and Lomonaco, V. and Bacciu, D.},
  booktitle={IEEE Conference on Computer Vision and Pattern Recognition Workshops (CVPRW)},
  year={2022},
  doi={10.1109/CVPRW56347.2022.00424}
}

@MISC{coumans2020pybullet,
  author = {Coumans, E. and Bai, Y.},
  title =  {PyBullet, a Python module for physics simulation for games, robotics and machine learning},
  year = {2020}
}

@article{raffin2021stablebaselines,
  author  = {Raffin, A. and Hill, A. and Gleave, A. and Kanervisto, A. and Ernestus, M. and Dormann, N.},
  title   = {Stable-Baselines3: Reliable Reinforcement Learning Implementations},
  journal = {Journal of Machine Learning Research},
  year    = {2021},
  volume  = {22},
  number  = {268},
  pages   = {},
  doi = {10.5555/3546258.3546526}
}

@misc{rusu2016policy,
  title = {Policy Distillation},
  author = {Rusu, A.A. and Colmenarejo, S.G. and Gulcehre, C. and Desjardins, G. and Kirkpatrick, J. and Pascanu, R. and Mnih, V. and Kavukcuoglu, K. and Hadsell, R.},
  year = {2016},
  journal = {arXiv},
  doi = {10.48550/arXiv.1511.06295},
}

@article{vandeven2019clscenarios,
  author = {{van de Ven}, G.M. and Tolias, A.S.},
  title = {Three Scenarios for Continual Learning},
  journal = {arXiv},
  year = {2019},
  doi = {10.48550/arXiv.1904.07734}
}

@article{lesort2020clrobotics,
  author = {Lesort, T. and Lomonaco, V. and Stoian, A. and Maltoni, D. and Filliat, D. and {D{\'i}az-Rodr{\'i}guez}, N.},
  title = {Continual Learning for Robotics: Definition, Framework, Learning Strategies, Opportunities and Challenges},
  journal = {Information Fusion},
  year = {2020},
  volume = {58},
  pages = {52--68},
  doi = {10.1016/j.inffus.2019.12.004}
}

@article{chaudhry2019reheasalcl,
  author = {Chaudhry, A. and Rohrbach, M. and Elhoseiny, M. and Ajanthan, T. and Dokania, P. and Torr, P.H.S. and Ranzato, M.},
  title = {On Tiny Episodic Memories in Continual Learning},
  journal = {arXiv},
  year = {2019},
  volume = {},
  pages = {},
  doi = {10.48550/arXiv.1902.10486}
}

@article{billard2019manipulation,
  author = {Billard, A. and Kragic, D.},
  title = {Trends and challenges in robot manipulation},
  journal = {Science},
  year = {2019},
  volume = {364},
  pages = {},
  doi = {10.1126/science.aat8414}
}

@article{yu2022manipulationRL,
  author = {Yu, C. and Wang, P.},
  title = {Dexterous Manipulation for Multi-Fingered Robotic Hands With Reinforcement Learning: A Review},
  journal = {Frontiers in Neurorobotics},
  year = {2022},
  volume = {16},
  pages = {},
  doi = {10.3389/fnbot.2022.861825}
}

@article{shintake2018softgrippers,
  author = {Shintake, J. and Caccucciolo, V. and Floreano, D. and Shea, H.},
  title = {Soft Robotic Grippers},
  journal = {Advanced Materials},
  year = {2018},
  volume = {30},
  pages = {},
  doi = {10.1002/adma.201707035}
}

@article{piazza2019hands,
  author = {Piazza, C. and Grioli, G. and Catalano, M.G. and Bicchi, A.},
  title = {A Century of Robotic Hands},
  journal = {Annual Review of Control, Robotics, and Autonomous Systems},
  year = {2019},
  volume = {2},
  pages = {},
  doi = {10.1146/annurev-control-060117-105003}
}

@article{villani2012controlhands,
  author = {Villani, L. and Ficuciello, F. and Lippiello, V. and Palli, G. and Ruggiero, F. and Siciliano, B.},
  title = {Grasping and Control of Multi-Fingered Hands},
  journal = {Advanced Bimanual Manipulation},
  year = {2012},
  volume = {80},
  pages = {},
  doi = {10.1007/978-3-642-29041-1_5}
}

@article{stachowsky2016slippage,
  author = {Stachowsky, M. and Hummel, T. and Moussa, M. and Abdullah, H.A.},
  title = {A Slip Detection and Correction Strategy for Precision Robot Grasping},
  journal = {IEEE/ASME Transactions on Mechatronics},
  year = {2016},
  volume = {21},
  number = {5},
  pages = {},
  doi = {10.1109/TMECH.2016.2551557}
}

@article{cossu2021clrnn,
  author = {Cossu, A. and Carta, A. and Lomonaco, V. and Bacciu, D.},
  title = {Continual learning for recurrent neural networks: An empirical evaluation},
  journal = {Neural Networks},
  year = {2021},
  volume = {143},
  number = {},
  pages = {},
  doi = {10.1016/j.neunet.2021.07.021}
}

@article{kroemer2021learningmanipulation,
  author = {Kroemer, O. and Niekum, S. and Konidaris, G.},
  title = {A review of robot learning for manipulation: challenges, representations, and algorithms},
  journal = {The Journal of Machine Learning Research},
  year = {2021},
  volume = {22},
  number = {1},
  pages = {},
  doi = {10.5555/3546258.3546288}
}

@article{laschi2023learningcontrol,
  author = {Laschi, C. and Thuruthel, T.G. and Lida, F. and Merzouki, R. and Falotico, E.},
  title = {Learning-Based Control Strategies for Soft Robots: Theory, Achievements, and Future Challenges},
  journal = {IEEE Control Systems Magazine},
  year = {2023},
  volume = {43},
  number = {3},
  pages = {},
  doi = {10.1109/MCS.2023.3253421}
}

@article{centurelli2022dynamic,
  author = {Centurelli, A. and Arleo, L. and Rizzo, A. and Tolu, S. and Laschi, C. and Falotico, E.},
  title = {Closed-Loop Dynamic Control of a Soft Manipulator Using Deep Reinforcement Learning},
  journal = {IEEE Robotics and Automation Letters},
  year = {2023},
  volume = {7},
  number = {2},
  pages = {},
  doi = {10.1109/LRA.2022.3146903}
}

@article{thuruthel2019rl,
  author = {Thuruthel, T.G. and Falotico, E. and Renda, F. and Laschi, C.},
  title = {Model-Based Reinforcement Learning for Closed-Loop Dynamic Control of Soft Robotic Manipulators},
  journal = {IEEE Transactions on Robotics},
  year = {2019},
  volume = {35},
  number = {1},
  pages = {},
  doi = {10.1109/TRO.2018.2878318}
}

@article{hinton2015distillation,
  author = {Hinton, G. and Vinyals, O. and Dean, J.},
  title = {Distilling the Knowledge in a Neural Network},
  journal = {arXiv},
  year = {2015},
  volume = {},
  number = {},
  pages = {},
  doi = {10.48550/arXiv.1503.02531}
}

@inproceedings{jayanthi2023distillation,
  title={DROID: Learning from Offline Heterogeneous Demonstrations via Reward-Policy Distillation},
  author={Jayanthi, S. and Chen, L. and Balabanska, N. and Duong, V. and Scarlatescu, E. and Ameperosa, E. and Zaidi, Z.H. and Martin, D. and Del Matto, T.K. and Ono, M. and Gombolay, M.},
  booktitle={2023 Conference on Robot Learning},
  doi={}
}

@article{traore2019discorl,
  author = {Traoré, R. and Caselles-Dupré, H. and Lesrot, T. and Sun, T. and Cai, G. and Diaz-Rodriguez, N. and Filiat, D.},
  title = {DisCoRL: Continual Reinforcement Learning via Policy Distillation},
  journal = {arXiv},
  year = {2019},
  volume = {},
  number = {},
  pages = {},
  doi = {10.48550/arXiv.1907.05855}
}

@article{pique2022clsoft,
  author = {Piquè, F. and Kalidindi, H.T. and Fruzzetti, L. and Laschi, C. and Menciassi, A. and Falotico, E.},
  title = {Controlling Soft Robotic Arms Using Continual Learning},
  journal = {IEEE Robotics and Automation Letters},
  year = {2022},
  volume = {7},
  number = {2},
  pages = {},
  doi = {10.1109/LRA.2022.3157369}
}

@inproceedings{donato2023plantreaching,
  title={Plant-inspired behavior-based controller to enable reaching in redundant continuum robot arms},
  author={Donato, E. and Ansari, Y.T. and Laschi, C. and Falotico, E.},
  booktitle={2023 IEEE RoboSoft},
  doi={10.1109/RoboSoft55895.2023.10122017}
}

@inproceedings{donato2022plantmovements,
  title={To Enabling Plant-like Movement Capabilities in Continuum Arms},
  author={Donato, E. and Ansari, Y.T. and Laschi, C. and Falotico, E.},
  booktitle={2022 I-RIM Conference},
  doi={10.5281/zenodo.7531338}
}

@article{puhlmann2022softhand,
  author = {Puhlmann, S. and Harris, J. and Brock, O.},
  title = {RBO Hand 3: A Platform for Soft Dexterous Manipulation},
  journal = {IEEE Transactions on Robotics},
  year = {2022},
  volume = {38},
  number = {6},
  pages = {},
  doi = {10.1109/TRO.2022.3156806}
}

@ARTICLE{Bianchi2023,
	author = {Bianchi, Diego and Antonelli, Michele Gabrio and Laschi, Cecilia and Sabatini, Angelo Maria and Falotico, Egidio},
	title = {SofToss: Learning to Throw Objects With a Soft Robot},
	year = {2023},
	journal = {IEEE Robotics and Automation Magazine},
	pages = {2–12},
	doi = {10.1109/MRA.2023.3310865},
	type = {Article},	
}

@ARTICLE{Nazeer2023,
	author = {Nazeer, Muhammad Sunny and Laschi, Cecilia and Falotico, Egidio},
	title = {Soft DAgger: Sample-Efficient Imitation Learning for Control of Soft Robots},
	year = {2023},
	journal = {Sensors},
	volume = {23},
	number = {19},
	doi = {10.3390/s23198278},
	}
    
    
    \appendix 
    \subsection{Glossary} \label{appendix:glossary}
        \textbf{Policy Learning}: The process by which a robot learns a set of rules or strategies (known as a policy) that dictate its actions in various situations. Policies can be learned using various techniques such as reinforcement learning, imitation learning, or a combination of both. In particular, in reinforcement learning, the robot learns by interacting with its environment, receiving feedback in the form of rewards or penalties based on its actions. Over time, it adjusts its policy to maximize the cumulative reward it receives.
        
        \textbf{Policy Distillation}: A technique used in reinforcement learning where knowledge from a complex policy (teacher) is transferred to a simpler policy (student). This is often done to reduce computational complexity or facilitate transfer learning between different tasks or environments.
        
        \textbf{Policy Demonstration}:  A machine learning approach where an agent learns a task by observing demonstrations provided by an expert. In the context of robot control, policy demonstration involves teaching a robot how to perform a task by showing it examples of desired behaviour rather than explicitly specifying a reward function or optimal policy.
        
        \textbf{Demonstration Sampling}: The process of selecting and collecting demonstrations or examples of behaviour to train a learning agent. Sampling techniques may involve random selection, prioritized sampling, or active learning strategies.
        
        \textbf{Knowledge Integration}: The process of incorporating new information or learned knowledge into an existing system or model. In the context of Continual Learning, knowledge integration involves updating or adapting the model based on new experiences or tasks.
\end{document}